\documentclass{elsarticle}
\pdfoutput=1
\usepackage{lineno,hyperref}
\usepackage{amsmath}
\usepackage{amssymb,amsthm}
\usepackage[inline]{enumitem}
\usepackage{float} 
\usepackage{comment} %
\usepackage{booktabs}

\usepackage{cleveref} %
\usepackage{mathtools} %

\usepackage{bm}
\usepackage{siunitx}
\usepackage{upgreek}

\usepackage{booktabs}

\renewcommand{\vec}[1]{\bm{\mathrm{#1}}}
\newcommand{\mat}[1]{\bm{\mathrm{\MakeUppercase{#1}}}}
\newcommand{\real}[1]{\mathbb{R}^{#1}}
\newcommand{\norm}[1]{\lVert #1 \rVert}
\newcommand{\nystrom}{Nystr{\"o}m}

\DeclareMathOperator{\rmse}{RMSE}
\DeclareMathOperator{\nrmse}{NRMSE}

\newcommand{\prmse}{\ensuremath{\gamma_{\mathrm{RMSE}}}}

\renewcommand{\ss}{\ensuremath{\mathtt{s}\rightarrow\mathtt{s}}}
\newcommand{\zms}{\ensuremath{\mathtt{zm}\rightarrow\mathtt{s}}}
\newcommand{\zmzm}{\ensuremath{\mathtt{zm}\rightarrow\mathtt{zm}}}

\modulolinenumbers[5]

\journal{Journal of \LaTeX\ Templates}

\begin{document}

\begin{frontmatter}
	
\title{Physics Informed %
Shallow Machine Learning %
for Wind Speed Prediction}

\author[1]{Daniele Lagomarsino-Oneto}%
\author[2]{Giacomo Meanti}
\author[3]{Nicol{\`o} Pagliana}
\author[2]{Alessandro Verri}
\author[1,4]{Andrea Mazzino}
\author[2,5,6]{Lorenzo Rosasco}
\author[1]{Agnese Seminara\corref{mycorrespondingauthor}}
\cortext[mycorrespondingauthor]{Corresponding author}
\ead{agnese.seminara@unige.it}

\address[1]{Department of Civil, Chemical and Environmental Engineering, University of Genoa, Genoa, Italy}
\address[2]{MaLGa, Department of Computer Science, Bioengineering, Robotics and Systems Engineering, University of Genoa, Genoa, Italy}
\address[3]{MaLGa, Department of Mathematics, University of Genoa, Genoa, Italy}
\address[4]{National Institute of Nuclear Physics, Genoa section, Genoa, Italy}
\address[5]{Center for Brains, Minds and Machines, MIT, Cambridge, MA, USA}
\address[6]{Italian Institute of Technology, Genoa, Italy}

\begin{abstract}
The ability to predict wind is crucial for both energy production and weather forecasting. Mechanistic models that form the basis of traditional forecasting perform poorly near the ground. In this paper, we take an alternative data-driven approach based on supervised learning. We analyze a massive dataset of wind measured from anemometers located at 10 m height in 32 locations in two central and north west regions of Italy (Abruzzo and Liguria). We train supervised learning algorithms using the past history of wind to predict its value at a future time (horizon). Using data from a single location and time horizon we compare systematically several algorithms where we vary the input/output variables, the memory of the input and the linear \emph{vs} non-linear learning model. We then compare performance of the best algorithms across all locations and forecasting horizons. We find that the optimal design as well as its performance vary with the location. We demonstrate that the presence of a reproducible diurnal cycle provides a rationale to understand this variation. We conclude with a systematic comparison with state of the art algorithms and show that, when the model is accurately designed, shallow algorithms are competitive with more complex deep architectures.
\end{abstract}


\end{frontmatter}

%\linenumbers

\section{Introduction}
The global consumption of energy produced from wind raised from about 87 TWh per year in the early 2000, to over 3500 TWh per year in 2019.
This relative growth rate of about $+4000\%$, as well as that from solar (about $+60000\%$), are order magnitudes larger than that from %
oil (approximately $+25\%$), nuclear ($-3\%$) and other more traditional renewables like hydroelectric power (approximately $+42\%$)~\cite{ritchie_energy_nodate,noauthor_bp_nodate,smil_energy_2017}.
Thus energy from wind %
will play an increasingly important role in the energy industry and in the global economy in the near future. %
The ability to predict wind is essential for maintenance of wind power plants as well as for energy markets relying on predictions of the energy power produced from wind. 
Moreover, reliable predictions of wind speed are valuable for private citizens as well as for public administrations
concerned with safety in the case of hazard scenarios.

Atmospheric forecasting and weather predictions have traditionally relied on numerical simulations of model equations based on physics. However, these mechanistic models, that form the basis of traditional forecasting, have poor performance close to the ground. Winds near the surface are affected by several processes that occur at spatial and temporal scales that are below the resolution of the numerical simulations. To account for these unresolved mechanisms, alternative approaches use machine learning and predict wind  speed close to the ground from time series of measured data. For a complete survey on time series techniques, independent of a particular application, see for example \cite{parmezan_evaluation_2019}. \\

Following this data-driven approach, several works have been carried out with a growing trend in the use of Deep Learning tools. Among deep architectures, Long-Short Term Memory (LSTM) Neural Networks\cite{lindemann_survey_2021} and its variants have received increasing attention due to their particular suitability to deal with sequential data like time series. In the context of wind speed prediction a large number of specific strategies have been developed. 
Many efforts are directed at designing methods to capture the multi-scale nature of atmospheric dynamics, where many decades of spatial scales are dynamically coupled in a highly nonlinear process. The pipeline of these algorithms may combine a multi-scale feature extraction stage with a following regression algorithm. Feature extraction may be accomplished through Wavelet Transforms\cite{li_multi-step_2019, yousuf_short-term_2021}, Singular Spectrum Analysis\cite{fu_composite_2020,liu_smart_2019}, Empirical Mode Decomposition\cite{fu_composite_2020, ruiz-aguilar_permutation_2021}. Other authors attempt to embed this sensitivity to multi-scale dynamics directly into the architecture of a neural network\cite{araya_multi-scale_2020}. 
Besides Deep Learning architectures, different algorithms have been developed based on Machine Learning models (for example kernel methods, Support Vector Regression \cite{de_mattos_neto_hybrid_2020} and Gaussian Processes \cite{zhang_gaussian_2016}) as well as classical statistical models like ARIMA and SARIMA\cite{bivona_stochastic_2011,yousuf_short-term_2021} and stochastic processes\cite{bivona_stochastic_2011,carpinone_markov_2015}. Furthermore several hybrid models that combine techniques from different families have been considered\cite{de_mattos_neto_hybrid_2020,camelo_innovative_2018,yousuf_short-term_2021}.\\

Forecasting methods for wind generally need to address 
its non-stationarity, i.e.~that the statistical distribution of wind speed may vary in time. 
The rolling or moving window approach is a widely adopted solution to tackle non stationarity and it consists in updating the model by periodically retraining the algorithm eliminating obsolete data and adding fresh information given by newly available data.
Although this technique proves crucial in certain applications like financial markets, there is no clear evidence in favor or  against this method for wind speed forecast. \\

Predictive models are also classified according to their forecast horizon, ranging from Very Short term (less than 1 hour), Short term (up to about 4 hours) to Medium term (up to 24 hours ahead) but also Long term predictions (more than 1 day). This latter subdivision is somewhat arbitrary and does not immediately connect to a notion of predictability, which may be better captured by other physical time scales (e.g.~the correlation time of wind speed) that typically change considerably with location. Moreover, the definition of ``Long Term'' as longer than one day is peculiar to data driven models, whereas physics-driven models typically forecast several days ahead. Note also that a forecast may be achieved by learning one specific model for the desired horizon, or by inferring directly 
an array of future values at different horizons either recursively or all together \cite{li_multi-step_2019,liu_smart_2019,li_smart_2019}.  %
Some works exploit information carried by other meteorological variables, like air pressure or temperature\cite{trebing_wind_2020}, or include spatial correlations among observations from different geographical locations in a network\cite{xu_multi-location_2022,zhu_wind_2018,messner_online_2019}. Remarkably, wind direction, which is usually available together with wind speed, has been rarely exploited to design features for wind speed forecast, with few exceptions\cite{trebing_wind_2020, chitsazan_wind_2017}. \\

Here, we analyze a massive experimental dataset of wind measured from anemometers located at 10 m height in 9 locations within the Abruzzo region in the central part of Italy and 23 locations in the Liguria region, in north western Italy. 
These two areas were chosen because of their complex orography and because of the interaction between land and sea circulations, which make wind prediction extremely challenging. We use these data to train supervised learning algorithms that use the past history of wind to predict its future values at different horizons. We first analyze a single location and a single time horizon and compare systematically several different algorithms where we vary: the input/output variables; the past history used for training; the linear \emph{vs} non-linear statistical model. 
Motivated by these results, we %
extend the analyses across all locations and all forecasting horizons.
We find that the optimal design as well as its performance can vary considerably with location; for example, the inclusion of wind direction improves performances in about half of the locations. Furthermore accounting for non-stationarity with a rolling window approach does not improve performance. 
We demonstrate that where and when the diurnal cycle is robust, the input data should include at least 24h of past history, to take advantage of the regularity of the pattern. This simple design principle is valid for all intermediate forecast horizons, that are most affected by the daily periodicity. 
Although the optimal algorithms vary with location, we identify a single model that preserves good performances across all datasets. 
By introducing a measure of performance relative to a widely used standard in atmospheric modeling, we demonstrate that this algorithm is competitive with more complex state-of-the-art algorithms. We further corroborate the result by applying our algorithm to datasets used in state-of-the-art literature and comparing the exact same diagnostic. 

In \cref{section_problem_and_data_driven} our data-driven approach to wind speed forecasting is described together with the machine learning algorithms and the datasets that we use.
In \Cref{section_main_results} we show and analyze our main results on the experiments.
In \Cref{section_rolling} we discuss the effect of accounting for non-stationarity with a rolling-window approach.
In \Cref{section_comparison} we compare our data driven approach to different methods that have been used within the context of wind speed forecasting.
In \Cref{section_conclusions} we discuss some final remarks and observations that follow from this work.

\section{Data driven models for wind forecast}\label{section_problem_and_data_driven}

In this section we describe the problem of wind speed forecasting and the data-driven approach we use to derive algorithmic solutions.
Each time series contains data of wind speed and direction recorded hourly from the start of 2015 to the end of 2019 (more details about the datasets can be found in \Cref{section_datasets}).
Each data point $\eta_t$ is described by a triplet:
\[
\eta_t=(s_t,m_t,z_t)
\]
where $s_t$ is the speed of the wind at time $t$; $m_t$ and $z_t$ are the meridional component and the zonal component of the wind respectively, such that $s_t = \sqrt{m_t^2 + z_t^2}$. \\%The two wind components are calculated from the wind direction $\theta_t$ as $m_t = s_t\sin{\theta_t}$ and $z_t = s_t\cos{\theta_t}$, where $\theta_t$ is zero along the West-East direction, it grows counterclockwise and is measured in radians.\\

\noindent Our goal is use these data to learn a model that predicts the wind speed $s_{t+h}$ at a future time $t+h$, where $h$ defines the forecast \textit{horizon}.\\

\begin{figure}[H]
	\centering
	\includegraphics[width=0.6\linewidth]{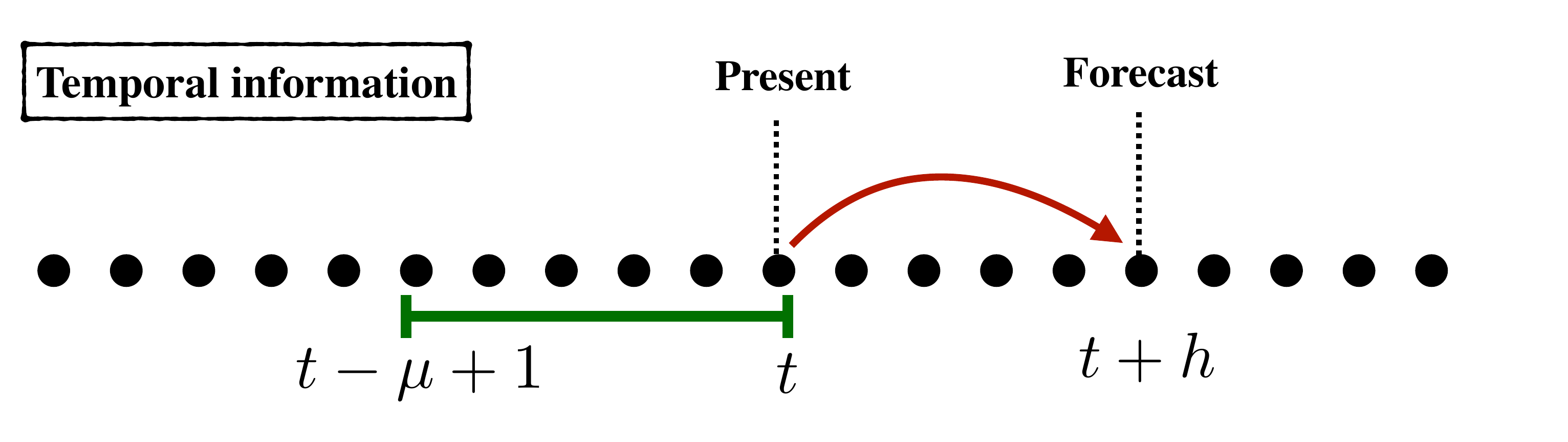}
	\caption{\textbf{Single sample definition} For each time $t$ in the time series we build an input vector from its past. The associated output is %
	the value of wind speed measured at $t+h$.} 
	\label{fig:sample_definition}
\end{figure}

\noindent We consider machine learning models, which infer the relation between the future value of the wind speed at time $t+h$ from the past $\mu$ measurements, where $\mu$ is called \textit{memory},
i.e. 
\[
\widehat{s}_{t+h}=\mathcal{F}(\eta_{t - \mu + 1}, \dots, \eta_t)
\]
where $\mathcal{F}$ denotes our machine learning model and $\widehat{s}_{t+h}$ our prediction at horizon $h$.
\Cref{fig:sample_definition} gives a pictorial representation of the wind speed prediction task.

\begin{figure}[H]
	\centering
	\includegraphics[width=0.7\linewidth]{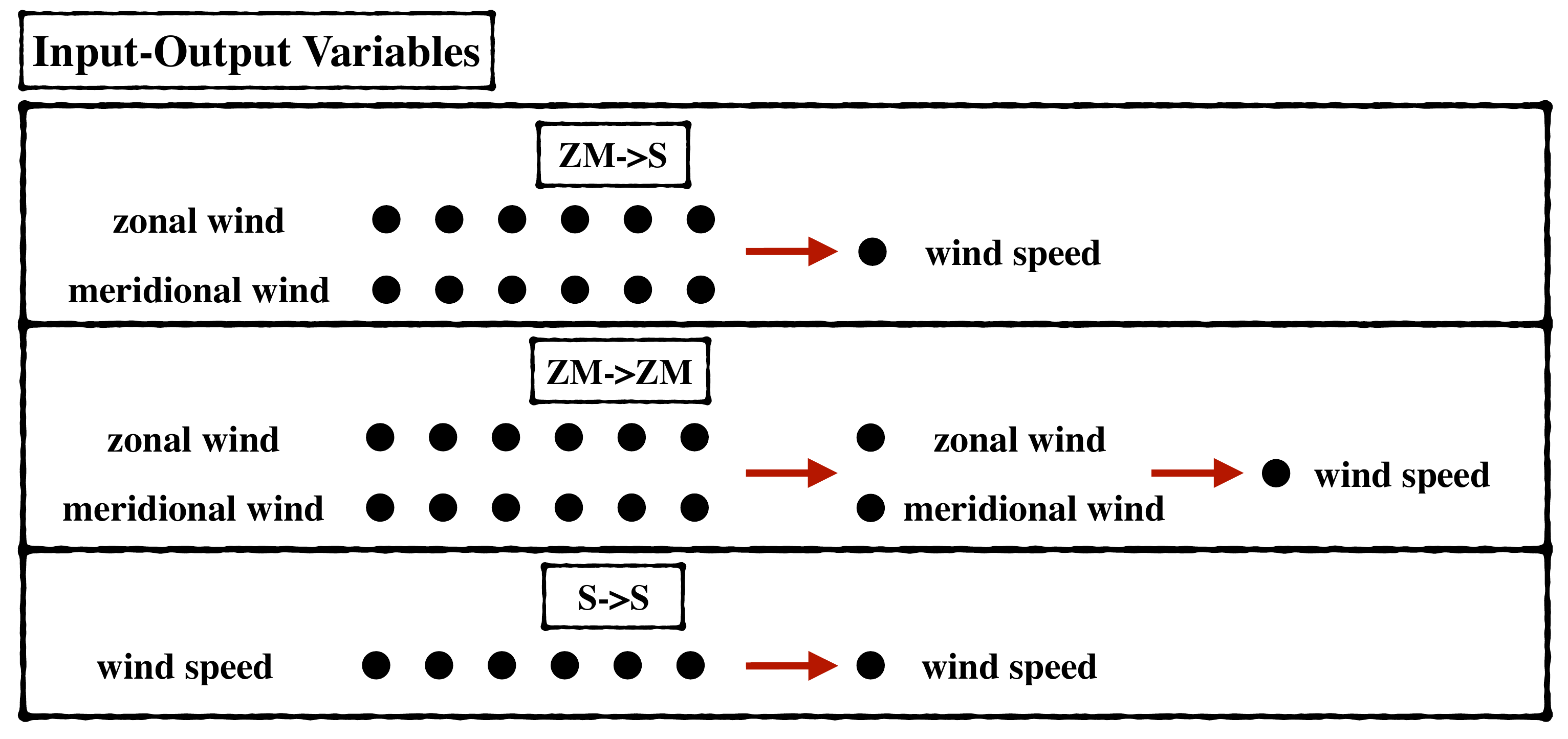}
	\caption{\textbf{Representation of the input/output setups used.} For each time $t$ in the time series we build an input vector from the past, either including both wind components (\zms~and~\zmzm) or considering only the wind speed (\ss). For \zms~and \ss, the associated output is the value of wind speed measured at $t+h$, where $h$ is the horizon. For \zmzm~we learn the two wind components at time $t+h$ separately and then reconstruct the wind speed.} 
	\label{fig:three_models}
\end{figure}

We study different combinations of horizon $h$, memory $\mu$, and input data $\eta_t$, in order to understand how they affect the overall prediction performance.
In particular we considered hourly horizons $h\in\{1,3,6,12,18,24\}$ and memories up to \num{3} days in the past $\mu \in \{2,6,24,48,72\}$.
For a fixed horizon $h$ and memory $\mu$ we consider the following options for designing the inputs and outputs (summarized in Figure~\ref{fig:three_models}).
\begin{description}
    \item[$\ss$] where both input and output are the wind speed:
    \[
    \widehat{s}_{t+h}=\mathcal{F}(s_{t - \mu + 1}, \dots, s_t)
    \,;
    \]
    \item[$\zms$] where the input is divided in zonal and meridional components and the output is wind speed
    \[
    \widehat{s}_{t+h}=\mathcal{F}((z_{t - \mu + 1}, m_{t - \mu + 1}),\dots,(z_t, m_t))
    \,;
    \]
    \item[$\zmzm$] where the input is divided in zonal and meridional components, each component of the wind vector is learned separately and the wind speed is computed from the components  
    \begin{align}
        \widehat{z}_{t+h}&=\mathcal{F}_1((z_{t - \mu + 1}, m_{t - \mu + 1}),\dots,(z_t, m_t))
        \\
        \widehat{m}_{t+h}&=\mathcal{F}_2((z_{t - \mu + 1}, m_{t - \mu + 1}),\dots,(z_t, m_t))
        \\
        \widehat{s}_{t+h}&=\sqrt{\widehat{z}_{t+h}^2+\widehat{m}_{t+h}^2}
        \,.
    \end{align}
\end{description}
More details about the computational procedures to derive the learning model $\mathcal{F}$ can be found in \Cref{section_supervised_learning}.

In order to measure the predictive performance of our models, we split the data from each location at a fixed date (January 1st, 2018). We then use all data before this date (the training set) to train model $\mathcal{F}$ and test the predictive performance of the model on the remaining data (test set). 
Note that data may be missing at specific dates for technical issues with the anemometers, and the missing data depend on location. Our splitting criterion leads to large variations in the training set size between stations%
; however it allows for a better comparison among the different sites by making the test sets uniform in size.
Throughout the paper, a \emph{static} approach to splitting is used where a model's training set is not updated in time. In Section~\ref{section_rolling} we motivate such choice with a case study to quantify the potential gain from updating the training set continuously with newly available samples.

We quantify the performance of the prediction with the normalised root mean squared error (NRMSE). 
For $n$ predictions it is defined as
\begin{equation}
\nrmse = \sqrt{\frac{\sum_{t=1}^n(s_t-\widehat s_t)^2}{\sum_{t=1}^n s_t^2}}.
\end{equation}

\subsection{Supervised learning}\label{section_supervised_learning}
For a given horizon $h$, memory $\mu$ and input-output for the model $\mathcal{F}$, we compose $n$ input-output pairs $(\vec{x}_t,y_t)_{t=1}^n$.
For example, for the $\zms$ model, pairs are defined as:
\begin{align}
    \vec{x}_t&=
    \big[
    z_{t - \mu + 1}, m_{t - \mu + 1},\dots,z_t, m_t
    \big]\in\real{d}
    \\
    y_t&=s_{t+h}
    \;
\end{align}
where $d = \mu\times k$ and $k$ is the number of variables in $\eta_t$.
Denote by $\mat{X}\in\real{n\times d}$ the lag matrix with rows $\{\vec{x}_t\}_{t=1}^n$, and $\vec{y}\in\real{n}$ be the vector of outputs with elements $y_t$.
Linear regression assumes that future wind behavior depends linearly on its past trends: it aims to find coefficients $\vec{\upbeta}\in\real{d}$ which minimize the error
\[
	\dfrac{1}{n}\sum_{i=1}^n \norm{\vec{x}_i \vec{\upbeta} - y_i}^2
	=
	\dfrac{1}{n} \norm{\mat{X} \vec{\upbeta} - \vec{y}}^2.
\]
The prediction on a new point $\vec{x}_\mathrm{new}$ is given by $\vec{x}_\mathrm{new}\vec{\upbeta}$.
Linear dependencies do not allow the model to account for complex interactions between past and future behavior of the wind. Kernel ridge regression (KRR) introduces a non-linear transformation of the features via the kernel function $k: \real{d}\times\real{d}\to\real{}$ which intuitively measures the similarity between two data-points. In our experiments we used the Gaussian kernel which is defined as $k(\vec{x}_i, \vec{x}_j) = e^{-\norm{\vec{x}_i - \vec{x}_j}^2 / (2\sigma^2)}$. The solution to the KRR problem yields an estimator $\widehat{f}$ which can be used for inference
\[ 
	\widehat{f}(\vec{x}_{\mathrm{new}}) = k(\vec{x}_{\mathrm{new}}, \mat{X})(\mat{K} + n\lambda \mathrm{I})^{-1} \vec{y} 
\]
where $\mat{K}\in\real{n\times n}$ is the kernel matrix with values $\mat{K}_{ij} = k(\vec{x}_i, \vec{x}_j)$ and $\lambda$ is a regularization parameter which ensures the problem is well-posed. %
The \nystrom{} method~\cite{williams_using_2001,smola_sparse_2000} is used to approximate the KRR solution maintaining good accuracy while greatly improving the algorithm's running time. 
The key of the reduction in computational complexity is to choose a small subset of $m \ll n$ points uniformly at random from the training-set $\mat{X}$ and approximating the kernel matrix $\mat{K}$ by a low-rank matrix depending on the $m$ selected training points.
We used the Falkon algorithm~\cite{rudi_falkon_2017,meanti_kernel_2020} which solves \nystrom{} KRR very efficiently by running a preconditioned conjugate gradient iteration, and can use GPU resources to further decrease the running time.

\subsection{Hyperparameter selection}\label{sec:hyperparams}
For each location under consideration we used five-fold cross-validation to estimate model hyperparameters (e.g~$\lambda$ and $\sigma$ for KRR). We used a two-step grid-search where in the first step we looked for the hyperparameters maximizing the $R^2$ score on a coarse grid, and in the second step we refined the grid around the optimum of the first step. The number of \nystrom{} centers $m$ were set to $10\cdot\sqrt{n}$ which provided a good trade-off in terms of accuracy versus time.

\subsection{Datasets}
\label{section_datasets}
\begin{figure}
    \centering
    \includegraphics[width=1.\textwidth]{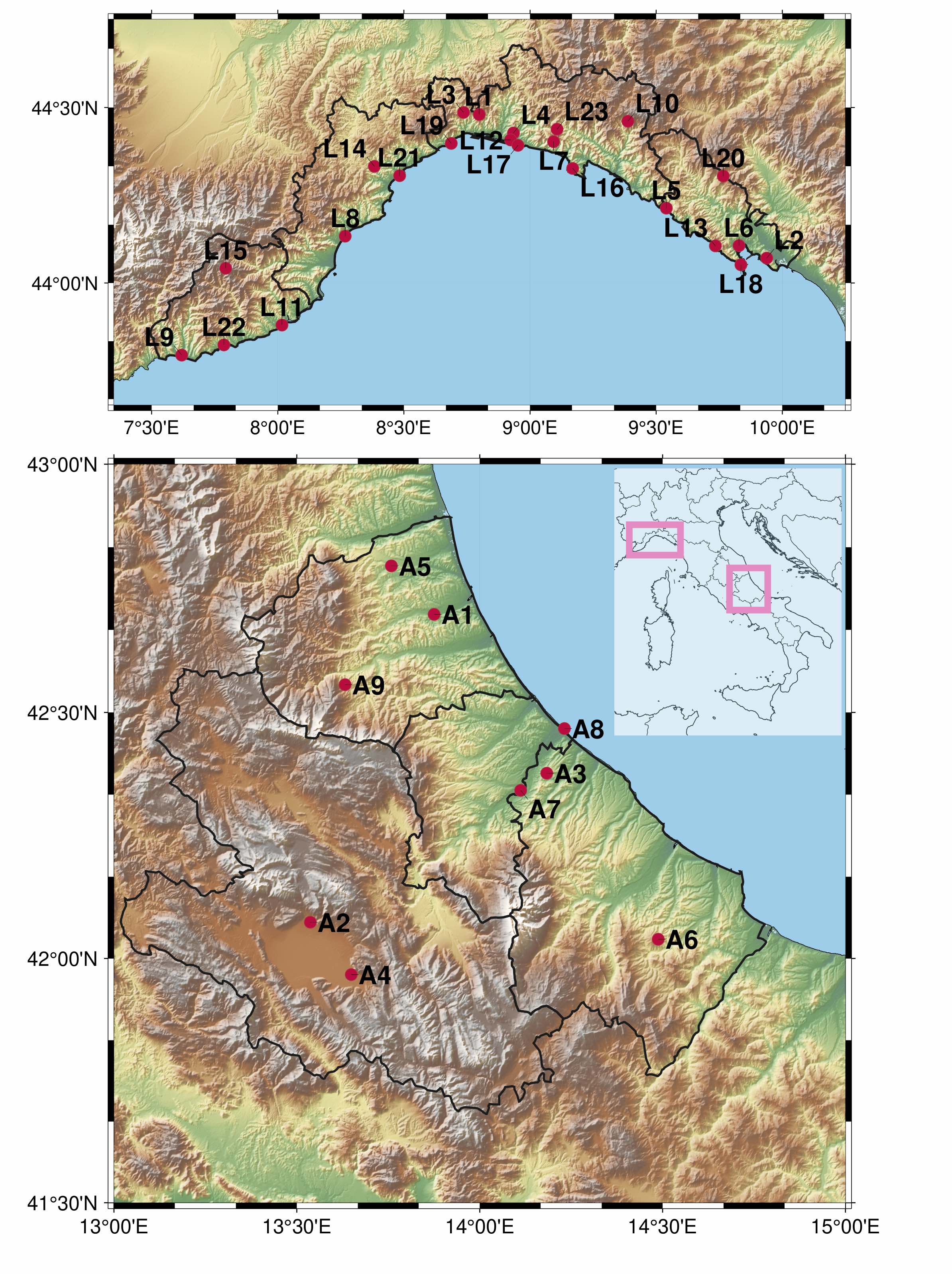}
    \caption{Physical location of the 32 anemometers in our dataset.}
    \label{fig:mappe}
\end{figure}

All time series analysed in our work consist of observations recorded by anemometers located $10$ meters above ground level. 
We take measurements from 32 meteorological stations in the Liguria and Abruzzo~\cite{noauthor_ufficio_nodate} regions of Italy, both characterized by the presence of a complex orography and proximity to the sea (see Figure~\ref{fig:mappe}).
The data series span a period between 4 and 7 years, depending on the station. 
Each location shows significantly different features affecting the wind speed predictability in different ways. 
Each raw time series is provided with a variable sample time of 10, 15 or 30 minutes for stations in Abruzzo and 1 hour for stations in Liguria. A sliding window average has been applied to obtain a uniform time step of one hour on the whole dataset.\\
Most recent literature uses data coming from wind farms, where the anemometers are typically around $90$ meters above ground. Note that within the atmospheric boundary layer, the vortical structures typical of turbulence (eddies) have a size that scales with distance from the ground. Therefore wind at \SI{90}{\meter} from the ground is more predictable as it changes over longer timescales, while wind at \SI{10}{\meter} from the ground (our datasets) is dominated by small eddies which change on a short timescale and are hard to predict.

\section{Results}\label{section_main_results}

In this section we describe the main results we obtained by running different models on the wind speed forecasting task. We begin by analysing the behavior of our models on two representative stations, comparing different inputs, outputs and model types. The observations made on this subset of locations are then taken into account to inform a full analysis of the whole set of stations.

\subsection{Analysis of two case studies} \label{preliminary_survey}
The stations considered here are A1 in Abruzzo and L1 in Liguria (see \Cref{fig:mappe}) as representatives of the whole dataset. They are geographically distant, and very different from a morphological point of view.
We fix the prediction task to wind speed prediction at a \num{3} hour horizon, and we wish to identify how the following parameters affect predictive performance.

\begin{enumerate}
    \item Input and output variables. We wish to determine which input-output variables out of $\ss$, $\zms$ and $\zmzm$ results in better predictive performance. $\ss$ consists in predicting future wind speed from past wind speed, $\zms$ uses both wind speed and direction in the input through the zonal and meridional components, and $\zmzm$ predicts both zonal and meridional components separately, to then reconstruct the wind speed itself.
    \item Model class. We aim to distinguish between the performance of \emph{linear} models (with the linear least squares (LLS) algorithm), and \emph{non-linear} models represented by KRR. 
    \item Memory $\mu$. The last crucial parameter is the amount of past data considered in each input point. For this parameter we experiment with values between \SI{2}{\hour} and \SI{72}{\hour}.
\end{enumerate}

\begin{figure}[H]
	\centering
	\includegraphics[width=0.95\linewidth]{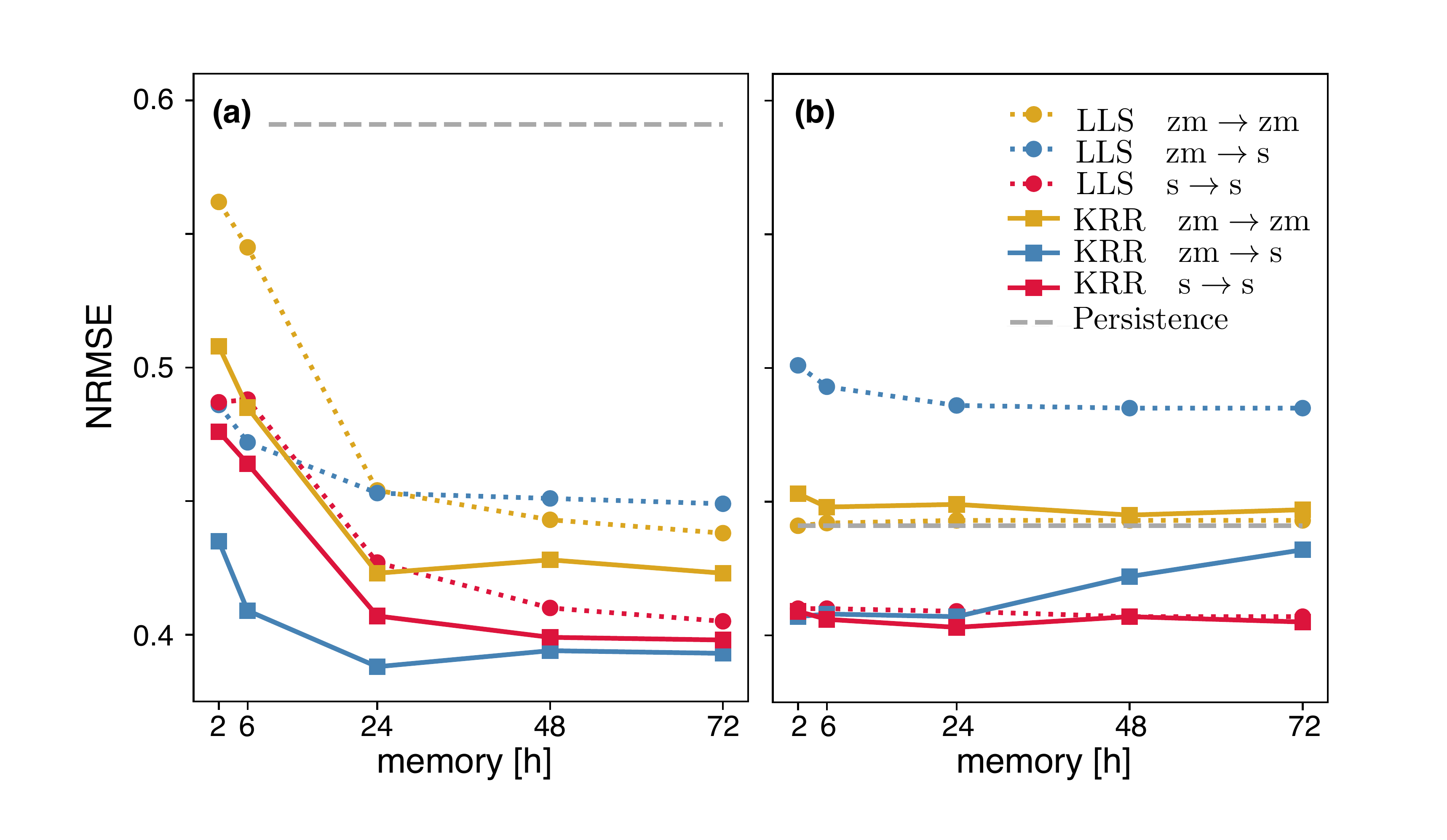}
	\caption{\textbf{Prediction accuracy for different memories on stations A1 and L1}. Panel (\textbf{a}) shows the predictive performance on station A1. All models improve considerably over the persistence (dashed grey line). The best non-linear model (solid lines, indicated with KRR) is the $\zms$ model (solid blue line) with \SI{24}{\hour} memory, which improves on the best linear model (dashed lines, indicated with LLS). Panel (\textbf{b}) displays results on station L1. From this other location a different picture emerges: the improvement over persistence is lower, and the best linear model is on par with the best non-linear one. All predictions were performed on a 3 hour horizon.}
	\label{fig:different_behaviors}
\end{figure}

The analysis of predictive accuracy as measured by the $\nrmse$ for the different parameters described above is shown in Figure~\ref{fig:different_behaviors}, and is compared to the performance of a na{\"i}ve model (the \emph{persistence} model), whose predictions for time $t + h$ are the observed values at time $t$. From panel (a) we make a few observations:
\begin{itemize}
    \item All models outperform the persistence model, reducing the $\nrmse$ by up to $20\%$ which is quite significant.
    \item The input design which performs best is the one which takes both wind components as inputs and predicts the wind speed directly, supporting the importance of wind direction for forecasting. However, restricting the comparison to the linear models, the $\ss$ input-output design performs better. Hence gains from using wind direction can only be leveraged by non-linear models, due to the non-linear dependency of wind speed from the two components of the wind vector. 
    \item For fixed input and output variables, non-linear algorithms are systematically better than their linear counterparts.
    \item Overall using \SI{24}{\hour} of memory seems to provide the best trade-off between performance and input size. Higher amounts of memory do not seem beneficial, and lower amounts of memory worsen performance noticeably.
\end{itemize}

To evaluate the robustness of these initial observations, we provide the same comparison for a second location (L1), see Figure~\ref{fig:different_behaviors}(b). %
First, improvements in $\nrmse$ over the persistence model drop dramatically to at most $4\%$. 
Second, the use of zonal and meridional components in the input brings no benefit (even for non-linear models). Third, linear and non-linear models achieve the same performance. Fourth, increasing memory above \SI{6}{\hour} provides no benefit. 
How can we interpret such discrepancies between two different stations? Is there a physical mechanism at the origin of these differences? In order to answer such questions we extend the analysis to a larger number of stations.

\subsection{Model design}

In this section we extend the survey of the different model types introduced previously (i.e.~taking into account input-output design, memory, model type) to the whole set of 32 stations. 
We further add into the mix of model parameters the forecast horizon. %
Short horizons are easier to predict even with simple models as their departure from the current state of the atmosphere is small. As the horizon increases, the chaotic dynamics in the atmosphere causes the wind to decorrelate from its current value more and more, thus the input bears less and less information about the output and predictions become more challenging. %
Overall we test a total of 186 model instances for each location, resulting in 5952 trained models. Noting that for each training, we also have to run the appropriate cross-validation to chose a model's hyperparameters (as described in \Cref{sec:hyperparams}), hence the total computational load very high. 
To reach the required scale in reasonable times, we rely on the Falkon library~\cite{rudi_falkon_2017, meanti_kernel_2020} which implements an approximation of KRR, coupled with clever optimization algorithms, on the GPU.

The aims of this extensive survey are to investigate \begin{enumerate*}[label=(\alph*)]
\item the role played by the input-output design and the model (i.e.~linear or non-linear),
\item the effects of memory on predictions, and their physical interpretation,
\item the possibility to identify a single model type which performs well in each geographical location.
\end{enumerate*}

\subsubsection{Input-Output Design}
\begin{figure}[H] 
	\centering
	\includegraphics[width=0.95\linewidth]{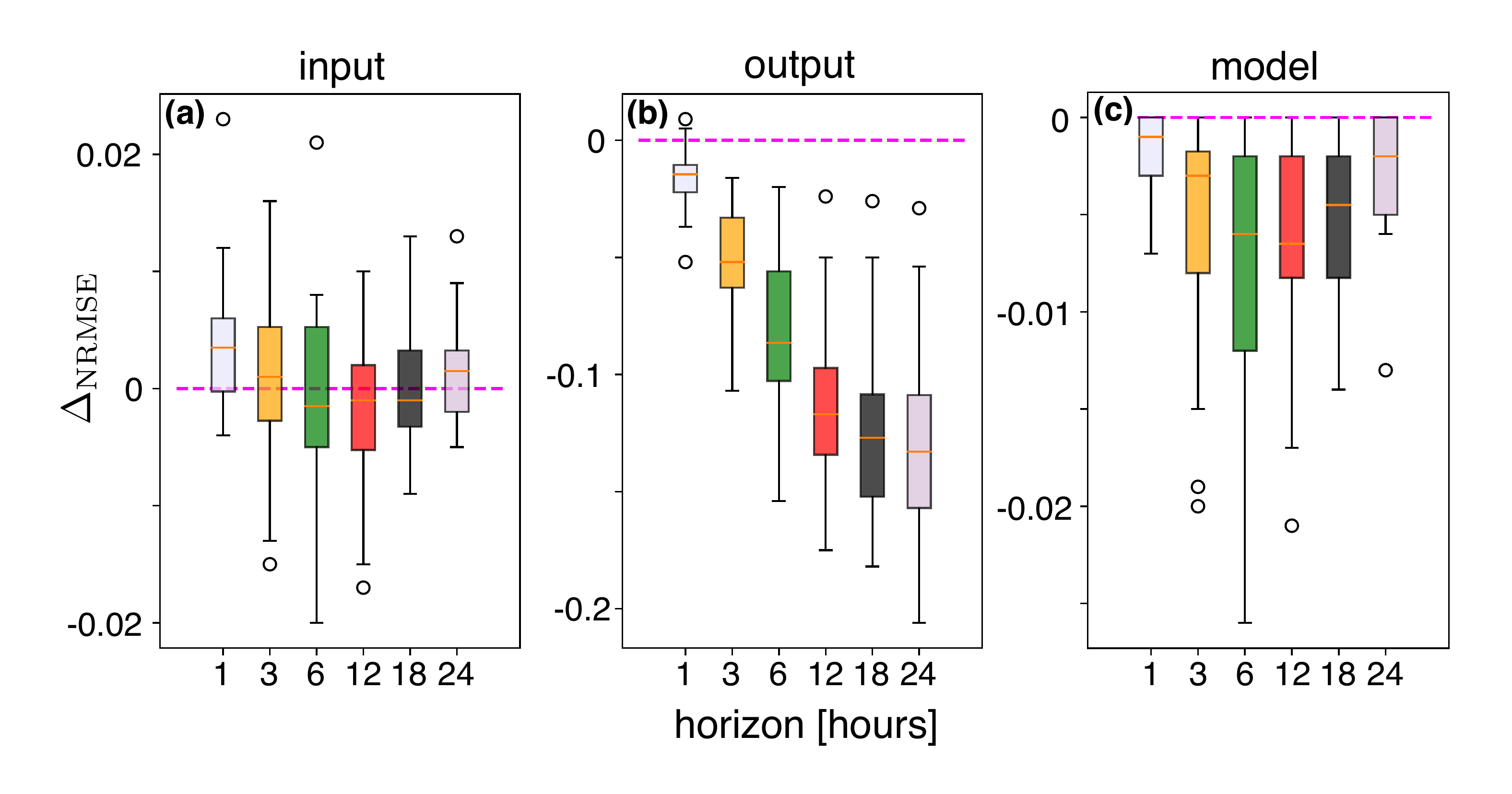}
	\caption{\textbf{Effect of different design choices on performance.} (\textbf{a}) Variation in the NRMSE ($\Delta_{\nrmse} = \nrmse(\zms) - \nrmse(\ss)$) when using both wind components \emph{vs} only wind speed in input (in both cases the output is wind speed). (\textbf{b}) Analogous variation in NRMSE ($\Delta_{\nrmse} = \nrmse(\zms) - \nrmse(\zmzm)$) when predicting wind speed \emph{vs} predicting separately the two wind components and then reconstructing the speed (in both cases the input includes both components of the wind). (\textbf{c}) Variation in NRMSE ($\Delta_{\nrmse} = \nrmse(\mathrm{nonlinear}) - \nrmse(\mathrm{linear})$) between the local best model and the best linear one, showing that linear models are competitive with non-linear models when we focus on next step forecast (1 hour ahead). In all panels memory $\mu$ corresponds to the optimal choice for each station and forecast horizons are indicated by colours and also reported on the x axis. 
	}
	\label{fig:IO_design}
\end{figure}

As can be seen in Figure~\ref{fig:IO_design} (\textbf{b}), we confirm the observations of the preliminary analysis: directly predicting the wind leads to much better performance, especially for long-term predictions.\\
On the input side, it can be observed in Figure~\ref{fig:IO_design} (\textbf{a}) that the benefits of including the direction in the input depend strongly on the location (note that this experiment used the KRR model). This result is consistent with the results of Section~\ref{preliminary_survey}, where two different stations had two different behaviors.
We can conclude that the influence of wind direction on its speed is complex, and not always helpful for improving predictions.\\
Finally, panel (\textbf{c}) shows that nonlinear models remain a better solution, with potentially moderate gains (e.g.~in the case of 6 hours ahead predictions), but virtually no downside. Another observation is that the performance improvement of KRR over LLS is smallest for horizons of 1 and 24 hours. This can be qualitatively understood by considering the predictability of wind at different forecast horizons.
A horizon of 1 hour is always within the correlation time of wind speed, hence for this horizon the time series can be described well by a linear autoregressive process. For 24 hours horizon, the effect of the diurnal cycle (easy to predict) becomes strong, and any variations on top of it are very hard to predict due to the long time scales. Therefore a more complex model has fewer advantages over a simpler one.

\subsubsection{The Role of Memory}

\begin{figure}[H]
	\centering
	\includegraphics[width=0.95\linewidth]{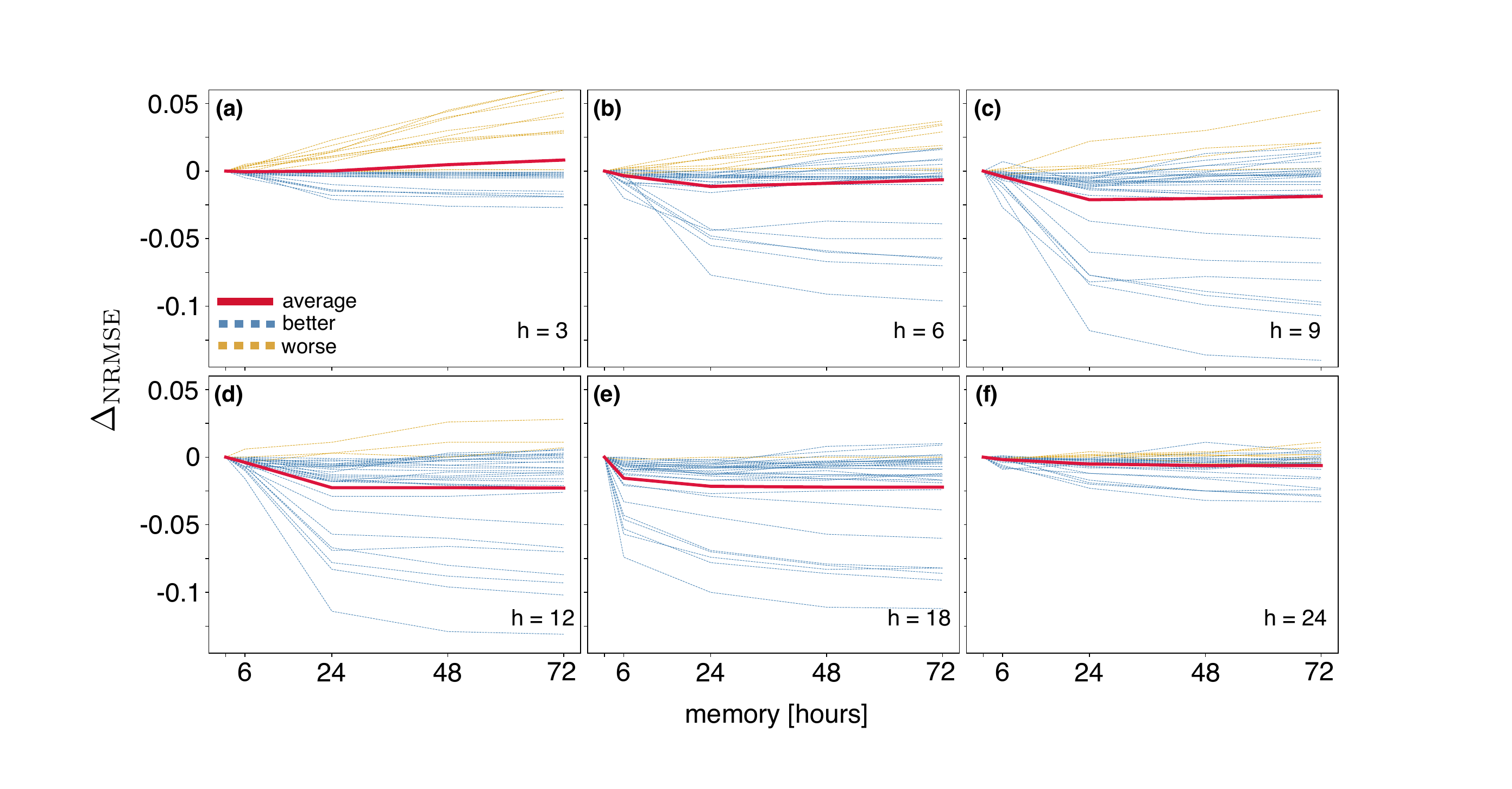}
	\caption{\textbf{Benefits of memory.} {\bf (a)}-{\bf (f)} Variation in performance $\Delta_{\nrmse} = \nrmse(\mu) - \nrmse(\SI{2}{\hour})$ between the locally best algorithm with memory $\mu$ and the same algorithm with reduced memory $\mu = 2$~hours for different time horizons ({\bf (a)} to {\bf (f)}  correspond to $ h=$~1, 3, 6, 8, 12, 24 hours). Each dashed line represents one location; blue and yellow mark locations where a memory of 24 hours improves and deteriorates performance respectively. Red solid lines represent averages over all locations.
At intermediate horizons (3 to 18 hours, panels (\textbf{b})-\textbf{(e)})), an input memory $\mu =$~24 hours is beneficial and longer memories lead to minor improvement.}
	\label{fig:memory1}
\end{figure}

We next ask what is the optimal amount of past memory to infer the future and how it depends on the forecast horizon.
In the preliminary analysis (Figure~\ref{fig:different_behaviors}) we observed that $\mu = \SI{24}{\hour}$ achieved the best performance whereas longer memory would increase the size of the input data without providing benefit for performance. 
To verify whether this result extends to other locations, we design the following experiment. For each location and horizon, we first identify the configuration (in terms of input-output, model type and amount of memory) with the lowest $\nrmse$. We refer to this configuration as the \textit{locally best model}.
We then take the locally best model's configuration (for input-output and model type), and train it with five different amounts of memory ($2, 6, 24, 48, 72$). In Figure~\ref{fig:memory1} we plot the difference between the $\nrmse$ with $\mu=2$ (short memory), and with the other values of $\mu$, for every station and horizon. Each line starts at zero (since $\Delta_{\nrmse} = \nrmse(\SI[parse-numbers = false]{\mu}{\hour}) - \nrmse(\SI{2}{\hour})$, and at the first point we have $\mu = 2$). For longer memories, it either decreases if longer  memory is beneficial, or it increases if longer memory is detrimental.

We find that memory affects our models' accuracy in a way that depends on the horizon. 
For short-term predictions (1 hour, Figure~\ref{fig:memory1}(a)), about 50\% of the stations are better predicted using a memory of 2 hours, rather than 24 hours. For most stations (78\%), the optimal memory increases when the horizon is set to $h=\SI{3}{\hour}$. The number of stations which benefit from longer memory further increases when the horizon is set to 6, 12 and 18 hours and for such medium term scenarios, the optimal memory is 24 hours for most stations (between 88\% and 97\%).
When the horizon is set to $h=\SI{24}{\hour}$, we observe that even though most stations (84\%) benefit from a longer memory, the improvement is marginal (an average decrease of 0.002 in $\nrmse$).
We note that when longer memory is beneficial, $\mu=\SI{24}{\hour}$ is a knee-point, i.e.~there is a considerable gain in switching from $\mu = \num{6}$ hours to $\mu = \num{24}$ hours, but further increasing memory to $\mu>24$ hours gives only small improvements. These modest gains come at a substantial computational price. %

We hypothesize that the role of memory laid out above can be traced back to the diurnal cycle in the atmosphere. In a nutshell, the diurnal cycle represents that many environmental quantities in the atmosphere undergo oscillations with a period of 24 hours, caused by periodicity of the sunlight. In the presence of a reliable diurnal cycle, the wind at time $t$ may be well-predicted by the wind at time $t-24$.
At very short horizons however, the wind changes little, thus better predictions may be achieved based on persistence, rather than by exploiting the diurnal cycle. In this case, 
a memory of 1 hour is optimal because a model which only takes the most recent data as input outperforms a model where the most informative data are combined with less informative data at previous times. At medium term, forecasts become more challenging as persistence is a poor predictor of wind. 
At these horizons, it is beneficial to include the full 24 hour cycle preceding the target time $t+h$, so that the prediction can benefit from the regularity of the wind.
Finally, predictions at $h=\SI{24}{\hour}$ are even more challenging, and all models incur in significant errors. In this case, the most recent data (at time $t$) is exactly $24$ hours before the target and is expected to be well correlated with wind at the target time. Moreover, there is a full diurnal cycle between the current time and the target time, thus including data prior to $t$ does not provide information about the most recent diurnal cycle, but about the preceding one. This is expected to be less informative, hence the marginal improvement in $\Delta_{\nrmse}$.

\subsubsection{Analysis of the Wind's Diurnal Cycle}

\begin{figure}[H]
	\centering
	\includegraphics[width=0.95\linewidth]{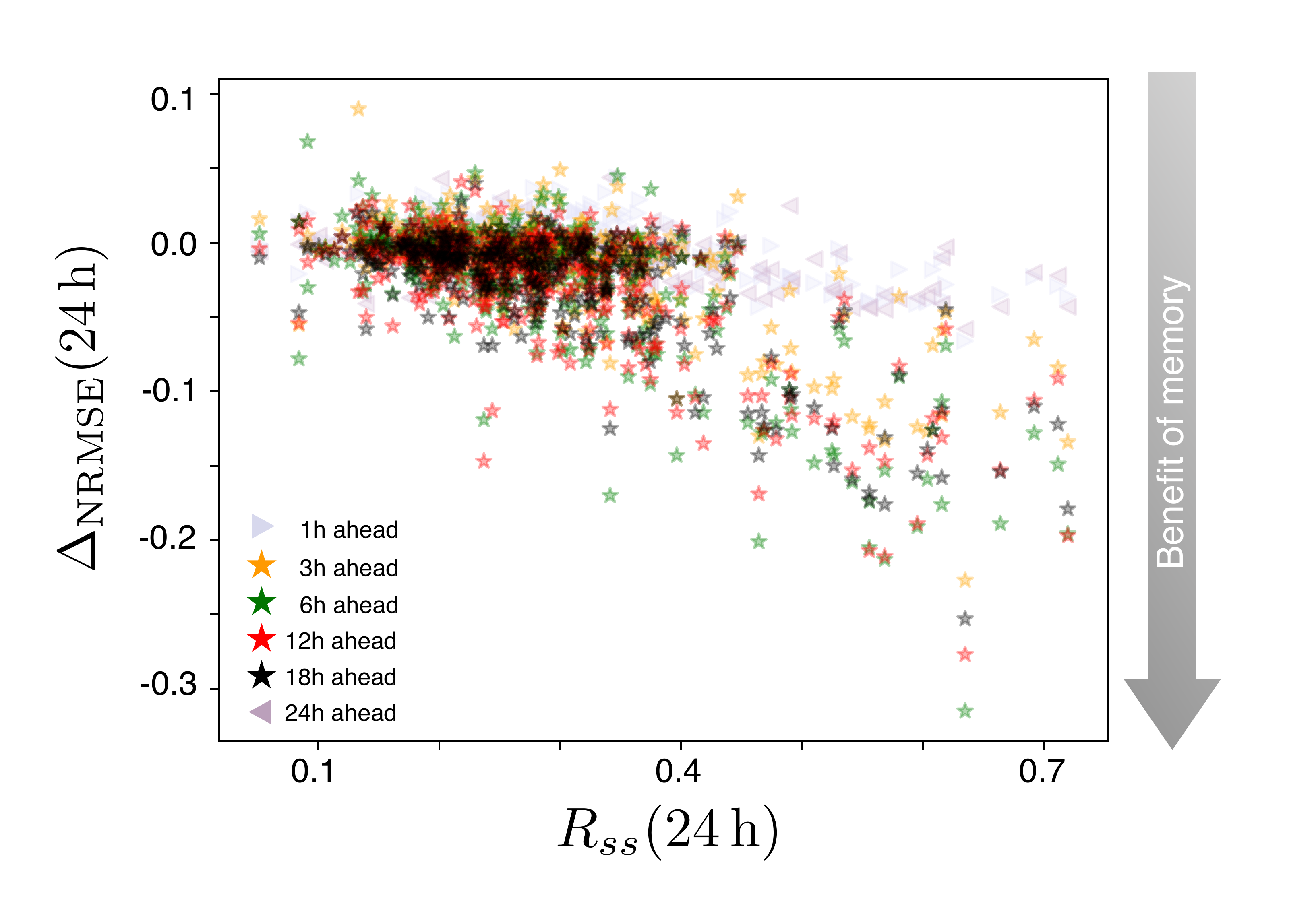}
	\caption{\textbf{At intermediate horizons, a 24-hour memory is beneficial in the presence of a strong diurnal cycle. %
	} Benefits of 24-hour memory are defined as $\Delta_{\nrmse} = \nrmse(\SI{24}{\hour}) - \nrmse(\SI{2}{\hour})$, i.e.~the difference in performance between the locally best algorithm with memory \SI{24}{\hour} and the same algorithm with reduced memory $\mu = 2$~hours. The strength of the diurnal cycle is defined as the normalized autocorrelation of the wind speed $s(t)$ at a lag of 24 hours: $R_{ss}(\SI{24}{\hour}) = \langle (s(t)-\bar s)(s(t+\mathrm{24 h})-\bar s)\rangle/\sigma^2_s$, where $\bar s$ and $\sigma_s$ are the mean value and the standard deviation of $s(t)$ respectively, computed over non-overlapping five-day windows. Each symbol represents data averaged over one month for a single location and forecast horizon (colored stars: intermediate horizons; grey triangles short and long horizons).}
	\label{fig:ac24_gain}
\end{figure}

To test the hypothesis that the diurnal cycle is at the origin of longer optimal memories for intermediate horizons, we proceed with a further analysis. 
We first %
compute the autocorrelation function of the wind speed time series and we keep track of the autocorrelation at 24 hours ($R_{ss}(\SI{24}{\hour})$) as a measure of the strength of diurnal cycle.
We then quantify the gain $\Delta_{\nrmse}$(24 h) provided by setting a memory $\mu = 24$ hours relative to  the choice of $\mu =  2$. %
Figure~\ref{fig:ac24_gain} includes all stations and forecast horizons and confirms the hypothesized relationship between the strength of the diurnal cycle $R_{ss}(\SI{24}{\hour})$ and the benefit of using a 24 h memory in the input. 
First, at intermediate forecasting times (colored stars), the major benefits of a 24 hour memory are clearly achieved when the diurnal cycle is stronger. 
Second, this clear trend vanishes at very short and long horizons ($h = 1$ hour and $h = 24$ hours, grey triangles).

\subsubsection{Selection of a single model} 

\begin{figure}[H]
	\centering
	\includegraphics[width=9 cm]{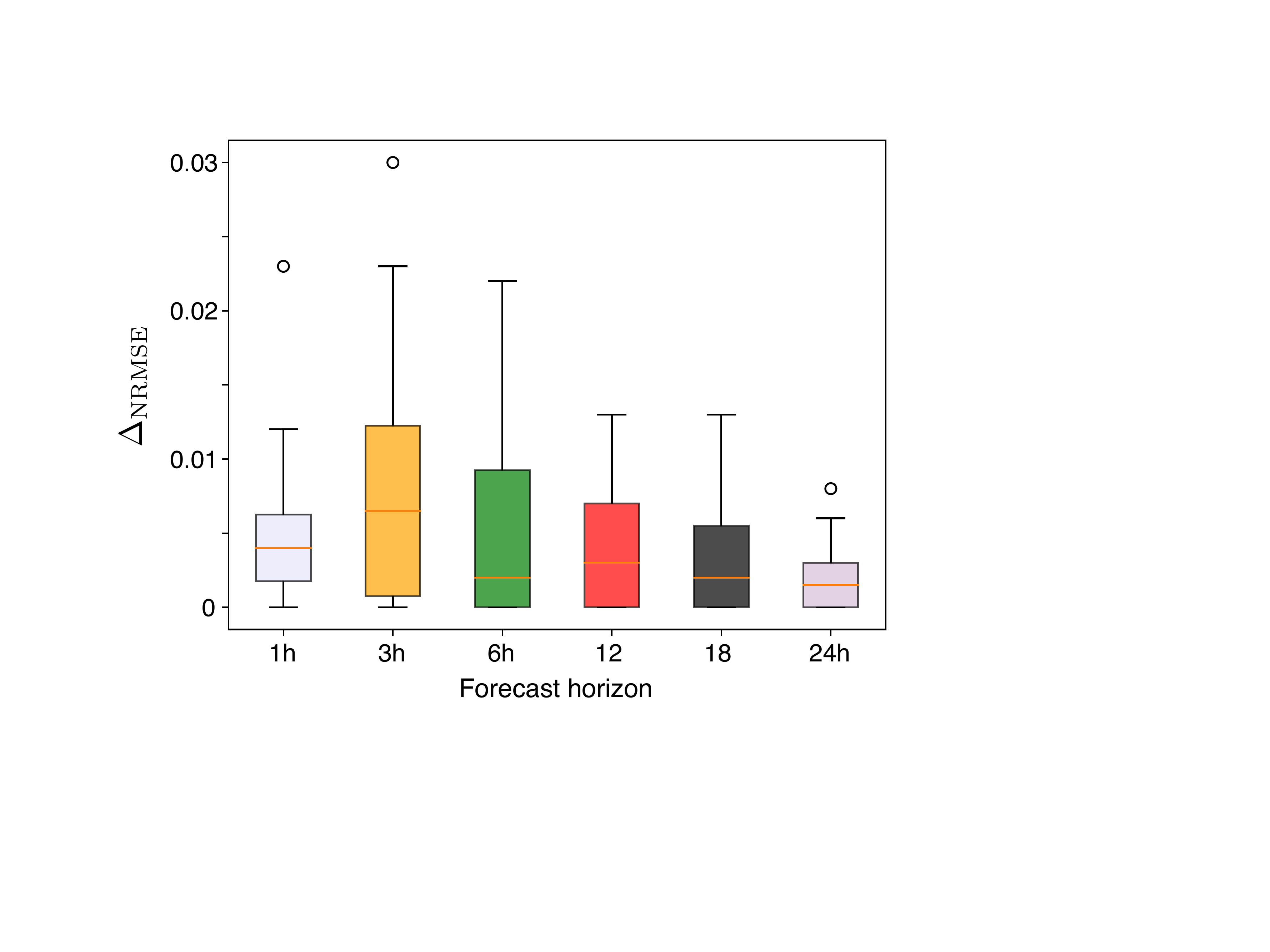}
	\caption{\textbf{Variation of performance between locally and globally best models.} Difference in $\nrmse$ ($\Delta_{\nrmse} = \nrmse(\mathrm{globally}) - \nrmse(\mathrm{locally})$) between locally best and globally best models as a function of the forecasting horizon. The distribution over the stations has an average which is always below \num{0.01} points (which correspond to a 1\% variation in the RMSE), and at most \num{0.03} points.}
	\label{fig:local_vs_uber}
\end{figure}

In the previous analyses, the locally best model was used, which changes for each station. However, it is desirable in practice to select a single model that may perform well over all stations so that it may be used as a default in the absence of better information. 
For each horizon, we call the model configuration which most frequently performs best the \textit{globally best model}. In Figure~\ref{fig:local_vs_uber} we show that the globally best model loses little accuracy over the locally best model (a few percentage points of $\nrmse$).
The globally best model features: $\mu = \SI{2}{\hour}$ for 1-hour-ahead predictions, $\mu = \SI{24}{\hour}$ for 3, 6, 12 and 24 hour ahead predictions and $\mu=\SI{72}{\hour}$ for 18 hour ahead predictions. The globally best input-output design is $\zms$ for all horizons except for the 18 hour horizon, where the global best is $\ss$.

\section{Non-stationary effects and rolling window approach
}\label{section_rolling}
The models described up to now, have been trained on all data before a cutoff date (01/01/2018), and tested on wind speed prediction with all available data after the same cutoff date.
This approach could incur in significant errors 
if the series is non-stationary and the statistical distribution of data changes over time. In this case, more recent data can better represent the evolution of the process while old data become obsolete.
To take into account the non-stationarity the model needs to be periodically updated~\cite{cheng_time_2015}. %
Several self-updating models have been used in different frameworks, including stock market forecasting~\cite{hota_time_2017,chou_forward_2018}, urban traffic control~\cite{mozaffari_vehicle_2015,haworth_local_2014}, ocean wave energy prediction~\cite{reikard_forecasting_2009} and streamflow estimation~\cite{lima_forecasting_2016}. 
The sliding window approach is a widely used updating technique that consists in periodically repeating training using a dataset from which obsolete data are removed and newly available data are added. 
The forgetting approach is a similar technique that consists in weighting the loss associated to the training examples to give more importance to recent data. %
Both approaches need to be retrained in order to include new information. 
In~\cite{messner_online_2019} the authors consider the problem of forecasting wind speed from data collected by different wind farms in France and Denmark. 
They consider a multi-valued linear lasso learning algorithm and update the model with a forgetting approach.
They show that updating the model improves performance with respect to the batch (non-updating) model, although the improvement may be somewhat marginal. %

Motivated by these results, we test whether performance of our model improves when we incorporate an updating procedure. For the sake of simplicity we consider a single station (A1) and a forecast horizon of 3 hours as a case study. We consider the locally best algorithm, which has a \SI{24}{\hour} memory, includes both wind components in the input and predicts wind speed. The update procedure of our model is based on a simple sliding window approach.
We consider the following 4 models (illustrated in Figure~\ref{fig:traintestsplit3}):
\begin{figure}[H]
	\centering
	\includegraphics[width=0.8\linewidth]{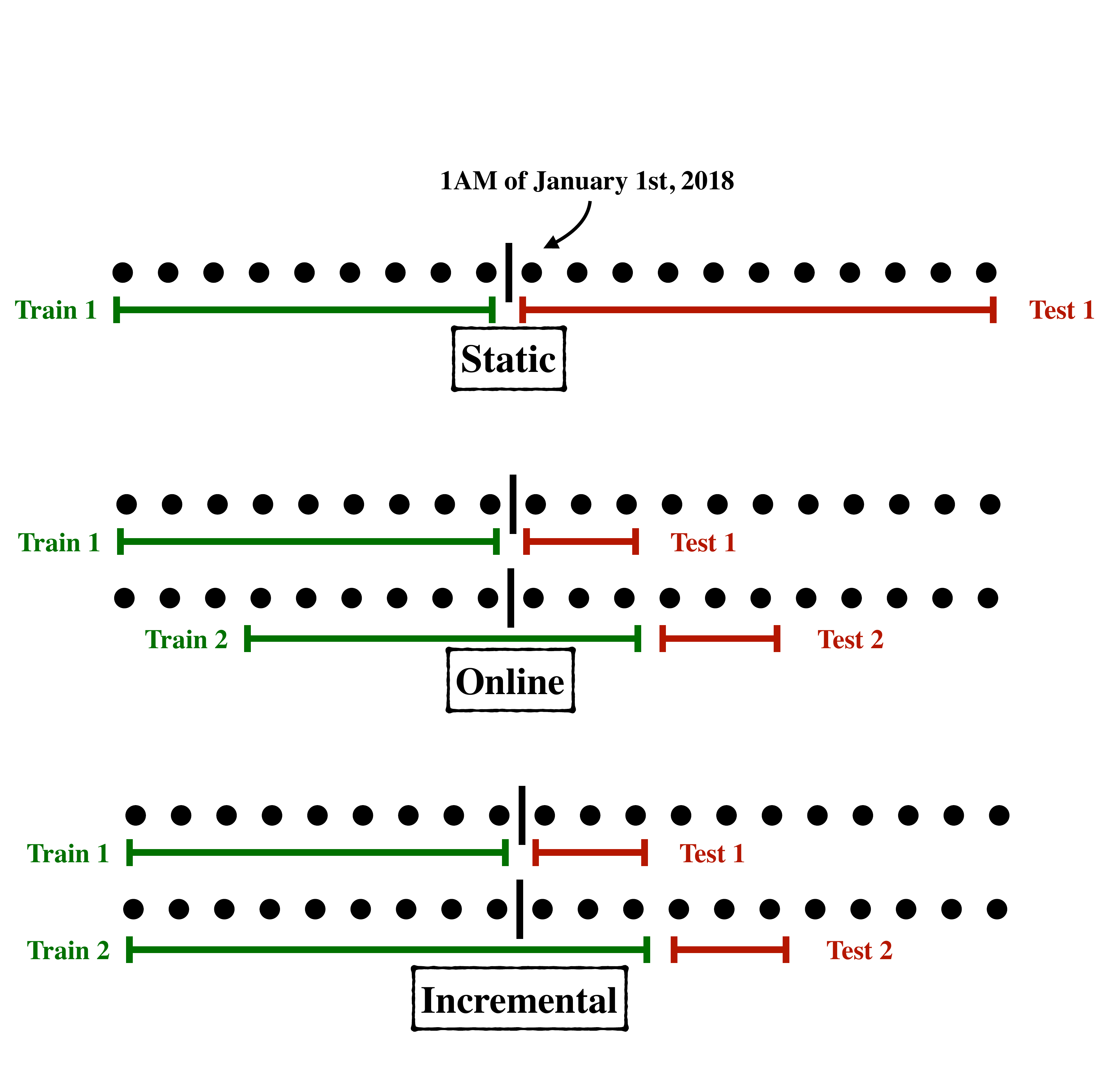}
	\caption{A graphical illustration of the updating processes of the static, online and incremental models.}
	\label{fig:traintestsplit3}
\end{figure}
\begin{itemize}
	\item The \textit{static} (non-updated) model is the one we discussed in the \textit{Results} section. It is trained only once, and uses a training set that starts from the beginning of the series until the last hour of the year 2017 (a total of $N=9805$ data points).
	\item The \textit{online} (updated) model, where starting from 1 AM of January 1$^{\text st}$, 2018, we train a model on the previous $N=9805$ data points and predict the data of the next week ($7\times 24 = 168$ samples). Then we re-train the model on the last $N$ available points before 1 AM of January 7$^{\text th}$ and test on the second week, repeating this procedure through the end of our dataset.
	We found that retraining more frequently than every $7$ days provided no advantage (data not shown).
	\item The \textit{incremental} (updated) model is similar to the \textit{online} model but when re-training, we consider all available data before the time we want to predict. Therefore the training set size increases as the testing dates move forward.
	\item The \textit{online (3m)} model is defined as the \textit{online} model but using a smaller training set with $2232$ samples ($3$ months) instead of the $9805$ used in the online model. This training set spanning one season should demonstrate the potential benefits of forgetting obsolete data.
\end{itemize}

\begin{figure}[H]
	\begin{center}
	\includegraphics[width=1.\linewidth]{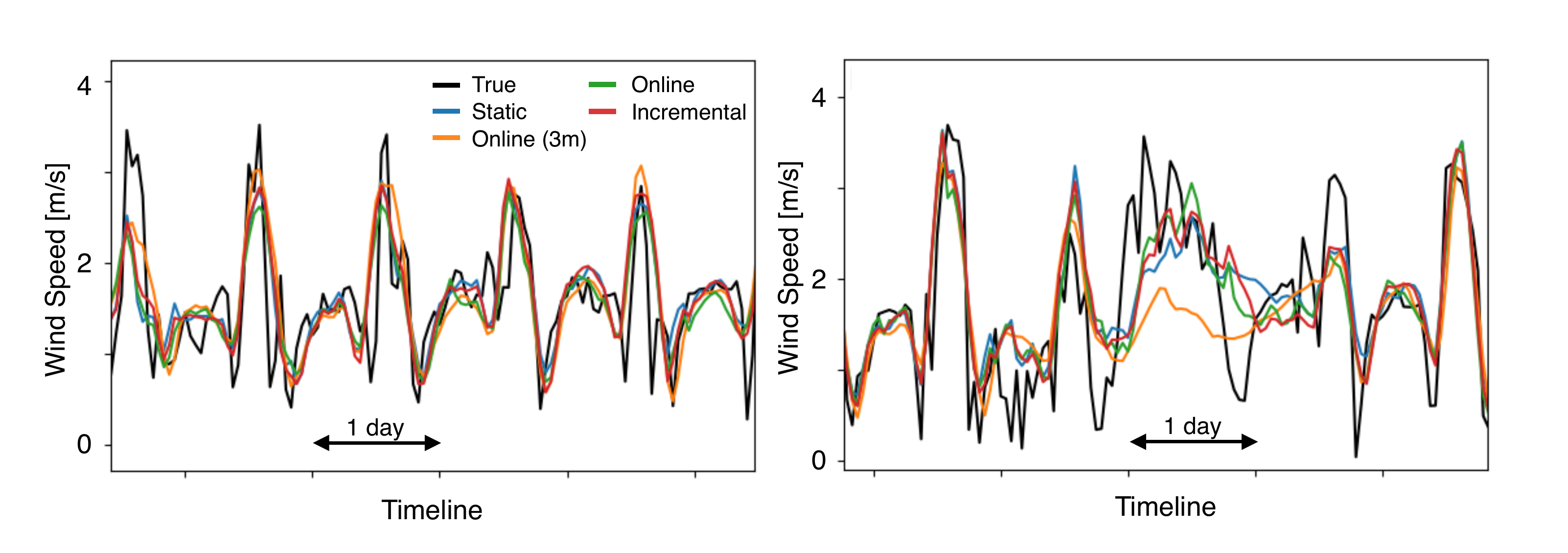}
	\end{center}
	\caption{Observed wind speed (black) compared to predictions obtained with the 4 different models described in the text (colored lines, see legend).}
	\label{fig:predictions}
\end{figure}

{\sisetup{round-mode=places,round-precision=3}
On the test set of the (A1) station, the $\nrmse$ of the four updating strategies described above is as follows. \emph{static}: \num{0.39045}, \emph{online}: \num{0.38182}, \emph{incremental}: \num{0.38123} and \emph{online (3m)}: \num{0.40121}. Hence the results are all quite close, with a few small differences: the incremental and online models perform best, followed by the static, and finally by the online model with just three months of history.
We argue that two effects are at play here: \textbf{a)} the benefits of updating the model with more recent information are visible from the difference between the \emph{online} and \emph{static} strategies (1\% in $\nrmse$). Such benefits are modest and they come at the cost of increasing the computational load noticeably since an expensive training step needs to be performed more frequently. The benefits are also not clearly noticeable in practice, as can be seen in \Cref{fig:predictions}, especially when the wind behaves regularly (left panel).
\textbf{b)} the benefits of a large training set can be seen from the difference between the \emph{online} and \emph{online (3m)} strategies (2\% difference in $\nrmse$). 
From \Cref{fig:predictions}(right panel) we can see an example of predicted time series where the \emph{online (3m)} model loses performance most clearly around an anomaly in the wind cycle. We propose that the smaller training set does not have enough statistical power to distinguish a variety of less common situations.
}

\section{Comparison with state-of-the-art models}\label{section_comparison}
\begin{figure}
    \centering
    \includegraphics[width=0.6\linewidth]{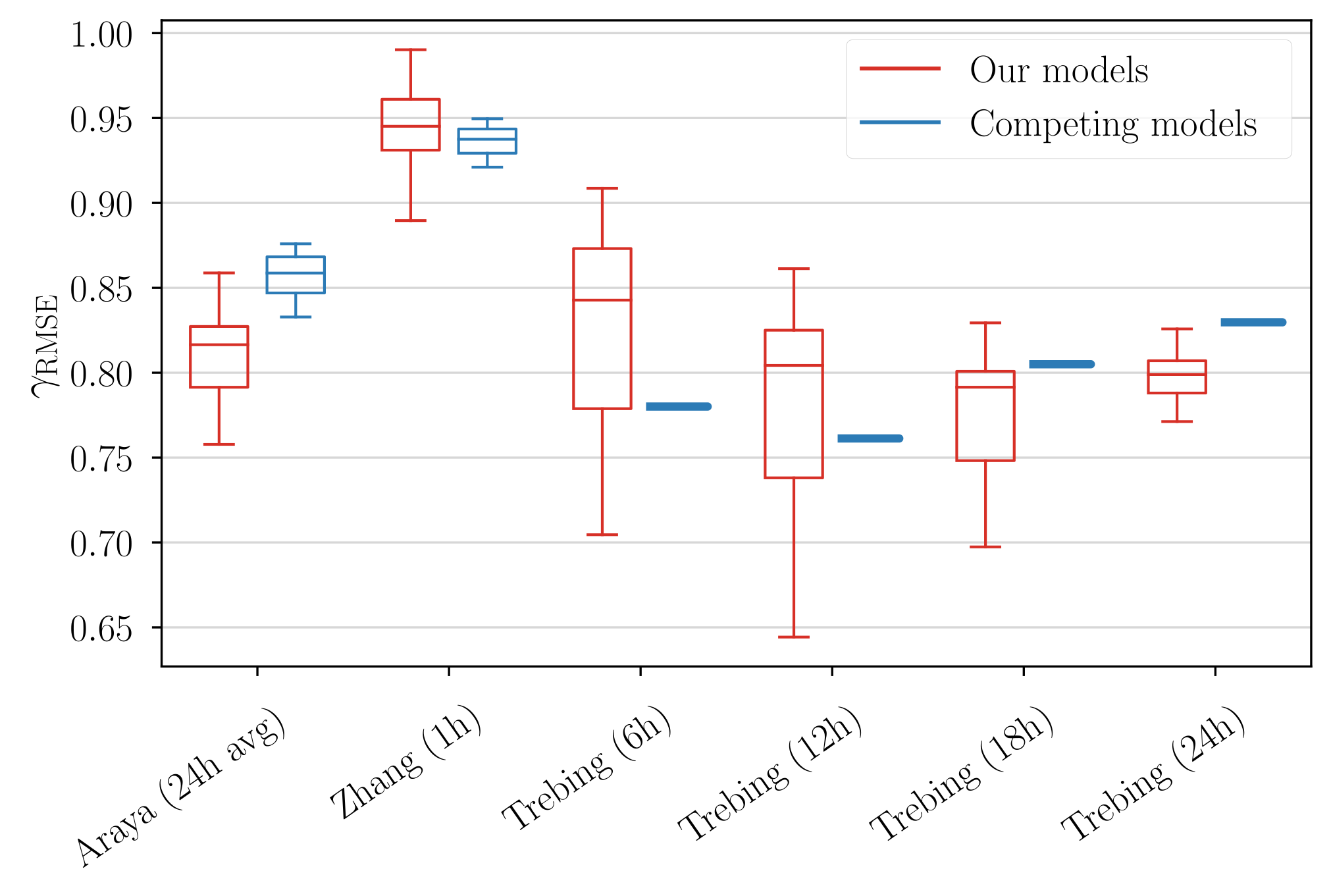}
    \caption{\textbf{Comparing improvement over persistence among different papers.} For our models (in red) we measured the $\prmse$ metric over all 32 stations for the same model: \nystrom{} KRR trained with the two hours of past data (memory) for the zonal and meridional wind components. For the competing models we took the best reported result for each site (even when obtained with different models).}
    \label{fig:persistence_comparison}
\end{figure}

In this section we compare the performance of our best models to that of other algorithms which have been proposed in the literature for predicting wind speed.
We first start with an approximate comparison that can be done by taking a third option,  tailored to the task of wind speed prediction. We compare two algorithms evaluated on two different datasets by tracking their respective improvements over the trivial prediction of the persistence model, applied to the relevant dataset.
To this end we define a metric $\prmse$ which captures the improvement of a specific algorithm ($\mathcal{A}$) over the persistence ($\mathrm{Pers}$) on a specific dataset in terms of RMSE
\[ \prmse(\mathcal{A}) = \dfrac{\mathrm{RMSE}(\mathcal{A})}{\mathrm{RMSE}(\mathrm{Pers})}. \]
The $\prmse$ metric can be used to compare algorithms evaluated on different datasets, as long as the error of the persistence is known. Nonetheless care must be taken to only compare results with the same forecasting horizon and sampling frequency, since the improvement over persistence strongly depends on this factor.
In Figure~\ref{fig:persistence_comparison} we use the $\prmse$ metric to compare our results with results from three papers that implement considerably more complex pipelines. 
Araya and others~\cite{araya_multi-scale_2020} designed a multi-scale deep learning model, based on the LSTM network architecture, with the aim of predicting hourly wind speed recorded at 20m high stations in four different sites in Chile. They report the averaged accuracy of 24h multi-step forecasts.
Zhang and others~\cite{zhang_gaussian_2016} used a Gaussian process stacked onto an autoregressive model, to predict wind speed at one step (hour) ahead for three different sites in China.
Finally, Trebing and Mehrkanoon~\cite{trebing_wind_2020} predict hourly wind speed in three danish cities at horizons of 6, 12, 18 and 24 hours. This latter work uses a convolutional neural network to blend both wind speed and other auxiliary data (i.e. temperature, pressure), from several cities at once.

For each comparison we compute a different error metric on our locations, to make it consistent with the compared work. We use the same \textit{globally best} model for all locations, while for the compared paper we take the best accuracy reported for each location or prediction horizon.
The results show that our model performs noticeably better than the multi-scale model of Araya and others.
The results of Zhang and others, are very close to ours, and in this case the $\prmse$ metric is higher since persistence becomes harder to improve upon for 1 hour ahead predictions.
Trebing and Mehrkanoon -- which only provide averages over three different sites -- use a model which is better than the proposed \nystrom{} kernel ridge regression on short term predictions (6 and 12 hours ahead), and worse on long term predictions (18 and 24 hours ahead). This suggests that the correlations between stations, and the auxiliary data, provide an advantage on the short term, but lose importance over long-term predictions where a model which takes purely the wind into account takes the lead. 

We then corroborate the comparison by training our models on publicly available datasets used in~\cite{araya_multi-scale_2020,trebing_wind_2020} adopting the same train/test splits and error metrics as in the original papers.
We select the \nystrom{} KRR model -- which on our datasets clearly outperforms linear models -- with the optimal memory as selected using 5-fold cross-validation. We test the two different input-output combinations which performed best on our data. The results for the optimal model are shown in Table~\ref{tbl:data_comparison}.
We confirm that our model achieves significantly better performance than Araya's original work, both when using the two wind components ($\zms$) and when using solely wind speed  ($\ss$). Comparing against Trebing's results we also recovered the trend suggested by the simple $\prmse$ metric: our model performs better on long-term and worse on short-term predictions. The results further indicate that the actual data, for example from multiple stations, using wind direction versus just wind speed, and auxiliary environmental measurements, plays a more important role than the model itself at improving the accuracy of wind speed prediction.

\begin{table}
	\centering
	\resizebox{.48\textwidth}{!}{
    \begin{tabular}{l p{1.2cm} p{1.9cm} p{1.9cm}}
		\toprule
		& Araya best & KRR($\mu$=24) $\zms$ & KRR($\mu$=24) $\ss$ \\
		\midrule
		e01   & 3.178  & \textbf{2.662}  & 2.702  \\
		b08   & 1.673  & \textbf{1.413}  & 1.488  \\
		d08   & 3.075  & \textbf{2.061}  & 2.081  \\
		d05a  & 2.406  & \textbf{2.135}  & 2.180  \\
		\bottomrule
	\end{tabular}}\hfill
	\resizebox{.48\textwidth}{!}{
	\begin{tabular}{l p{1.2cm} p{1.9cm} p{2.0cm}}
		\toprule
		& Trebing best & KRR($\mu$=48) $\zms$ & KRR($\mu$=48) $\ss$ \\
		\midrule
		6h   & \textbf{1.675}  & 1.814  & 1.714  \\
		12h  & 2.144  & 2.205  & \textbf{2.092}  \\
		18h  & 2.375  & 2.317  & \textbf{2.244}  \\
		24h  & 2.463  & 2.369  & \textbf{2.326}  \\
		\bottomrule
	\end{tabular}}
	\caption{\textbf{Comparing performances on available datasets used in recent works.} $\rmse$ of our models on the datasets of Araya and others~\cite{araya_multi-scale_2020} and Trebing, Mehrkanoon~\cite{trebing_wind_2020}. We used \nystrom{} KRR, with 24 and 48 hours of memory, and trained with different input variables. $\mathtt{s},\mathtt{z},\mathtt{m}$ indicate wind speed, zonal and meridional components and $\mathrm{aux}$ indicates auxiliary data from~\cite{trebing_wind_2020}.}
	\label{tbl:data_comparison}
\end{table}

\section{Discussion and conclusions}
\label{section_conclusions}
In this work, we develop a machine learning approach to predict wind at a future time purely from data, i.e.~with no aid from mechanistic modeling. %
 We conduct a systematic model selection through all our datasets, providing physical principles to understand the patterns that we observe and finally we propose a thorough comparison with state of the art algorithms. 
First, we compare models where both wind components \emph{vs} only wind speed are included in the input/output. This analysis quantifies the role of wind direction, which is a natural variable to be analyzed because anemometers typically record it together with the wind speed although this aspect is often neglected in the literature. We find that predicting wind speed from its two components is favorable in some locations but not in others. This result can be understood by noting that the dynamics of the atmosphere near the ground (where we focus our analysis) is particularly affected by the local orography and the features of the soil near the point of interest. Thus depending on the location, wind may interact with surface elements differently depending on its direction, explaining why it may be beneficial to include direction in the input. However this entirely depends on the details near the location of interest, hence the strong dependence on location. Predicting the two components of the wind separately and then reconstructing the speed from its components is never useful, perhaps because it involves two different models, leading to error buildup. 

Second, we analyze the role of memory, defined as the length of time used as input. We find that a memory of 24 hours is optimal for all intermediate horizons. We propose that this observation can be explained by the presence of regular diurnal cycles in wind speed, which are often observed in the atmosphere. To corroborate this intuition we quantify the strength of the diurnal cycle and find that it correlates strongly with the benefits of a 24-hour memory. These arguments  apply to intermediate horizons only because short and long horizons (here 1 hour and 24 hours) are either too easy or too hard to predict. One hour from now, the wind will not vary considerably thus persistence is a good predictor. At 24 hours from now, wind will be hard to predict, and the past history provides little information. These arguments are corroborated by the observation that non-linear models are clearly beneficial at intermediate horizons and only marginally at short and long horizons. 
We find that the best model in each location does not provide significant gains over using the single model that is most often selected as the best (typically the gains are below 1\% in NRMSE). Similarly we find little gain in taking into account non-stationary effects.

Third, we seek to compare our algorithm to the state of the art. We run into a major stumbling block due to the lack of benchmarking datasets: each manuscript tests a proposed algorithms on its own dataset -- whether public or proprietary. Clearly, to establish a fair comparison, different algorithms must be run on the same data. Hence to compare the models proposed in this paper with competing models we could either  \textbf{a)}~run the competing algorithm on our Liguria and Abruzzo datasets, or \textbf{b)}~run our proposed algorithm on the datasets used in the competing work. The first option requires to re-implement competing algorithms, which is time consuming and may result in artificial differences due to technical variations in the software. 
Hence we choose the second option, although it limits the possible comparisons to those works where data are made publicly available. 

The main conclusion is that when we run our model on the dataset published in two of the most recent papers, our performance is competitive with the state of the art algorithms.  
This is somewhat surprising because recent papers make extensive use of deep architectures and complex pipelines, sometimes enriching the input with additional variables. On the contrary, our accelerated kernel ridge regression models are considerably simpler to implement and additionally benefit from sound theoretical guarantees. 
This equivalence suggests that carefully optimizing the design of simple architectures paying attention to the input and output may be a fruitful alternative to the development of complex architectures that are typically less explainable.

Predictions of wind speed close to ground remain challenging both for physics-based models and for purely data-driven strategies. Mechanistic models pose conceptual as well as computational challenges. Atmospheric turbulence couples many spatial scales, hence numerical solutions of the equation of motion require a massive number of grid cells. Moreover, un-modeled mechanisms may affect the solutions. This is especially true near the ground, where interaction with the orography and the local elements on the ground cannot be modeled in detail. Data-driven approaches may recover some of these un-modeled effects statistically, however their performance is limited by the lack of information from the physical processes that take place in the atmosphere at different locations. 
Some of these information may be recovered by using spatio-temporal wind data as input, an approach that appears promising from current literature~\cite{trebing_wind_2020,zhu_wind_2018, xu_multi-location_2022, messner_online_2019}. 
More systematic hybrid approaches based on data assimilation techniques hold the promise to achieve the ideal merge of a mechanistic and a data-driven approach.  

\section*{Acknowledgements}

This work received support from: the European Research Council (ERC) under the European Union’s Horizon 2020 research and innovation programme (grant agreement No. 101002724 RIDING to A.S. and grant agreement No. 819789 SLING to L.R.); the Air Force Office of Scientific Research (AFOSR) under award number FA8655-20-1-7028, FA9550-18-1-7009, FA9550-17-1-0390 and BAA-AFRL-AFOSR-2016-0007 (European Office of Aerospace Research and Development) to L.R.; the National Institutes of Health (NIH) under award number R01DC018789 to A.S.; the EU H2020-MSCA-RISE project NoMADS - DLV-777826, and the Center for Brains, Minds and Machines (CBMM), funded by NSF STC award CCF-1231216 to L.R.
We thank the  ``Ufficio Idrografico e Mareografico'' of the Abruzzo Region and ARPAL for providing us with station wind data. Discussions with Federico Cassola are also warmly acknowledged.

\bibliography{Sections/references}

\end{document}